# PerceptionNet: A Deep Convolutional Neural Network for Late Sensor Fusion

Panagiotis Kasnesis, Charalampos Z. Patrikakis
Department of Electrical and Electronics Engineering
University of West Attica
Athens, Greece
pkasnesis@yahoo.gr, bpatr@puas.gr

Iakovos S. Venieris
School of Electrical and Computer Engineering
National Technical University of Athens
Athens, Greece
venieris@cs.ntua.gr

*Abstract*— **Human Activity Recognition (HAR) based on motion sensors has drawn a lot of attention over the last few years, since perceiving the human status enables context-aware applications to adapt their services on users' needs. However, motion sensor fusion and feature extraction have not reached their full potentials, remaining still an open issue. In this paper, we introduce PerceptionNet, a deep Convolutional Neural Network (CNN) that applies a late 2D convolution to multimodal time-series sensor data, in order to extract automatically efficient features for HAR. We evaluate our approach on two public available HAR datasets to demonstrate that the proposed model fuses effectively multimodal sensors and improves the performance of HAR. In particular, PerceptionNet surpasses the performance of state-of-the-art HAR methods based on: (i) features extracted from humans, (ii) deep CNNs exploiting early fusion approaches, and (iii) Long Short-Term Memory (LSTM), by an average accuracy of more than 3%.**

*Keywords*— **Convolutional neural network; deep learning; feature learning; human activity recognition; sensor fusion**

## I. INTRODUCTION

The proliferation of the Internet of Things (IoT) over the last few years, has contributed to the collection of huge amounts of time-series data. An IoT device with high sampling rates, such as a wearable, produces hundreds of data every second, resulting to a data explosion, considering the vast number of such devices connected over the internet. Through real-time or batch data processing, meaningful information is extracted, revealing daily patterns of individual owners or social groups. This information can be exploited by context-aware applications in order to enhance wellbeing [1], health status [2], facilitate smart environments [3] and improve security [4][5].

Context-aware applications are capable of discovering and reacting to changes in the environment their user is situated in [6][7]. Their effectiveness depends on four entities [8]: identity, location, time, and status (activity). The first three entities can be easily tracked without the need of deploying sophisticated algorithms. Extracting knowledge from sensors to define the human activity however, is a complex task, where the use of signal processing is mandatory.

Conventional signal processing techniques in HAR, apply mathematical, statistical or heuristic functions over raw motion data, in order to extract valuable features, identified as hand-crafted features (HCFs). Concretely, D. Figo et al. [9] categorize the feature extraction techniques in three domains:

the *time* domain (e.g., mean, max, mean values), the *frequency* domain (e.g., Fast Fourier Transformation) and *discrete representation* domain (e.g., Euclidean-based distances). These hand-crafted features, feed a classification algorithm, which after a training phase, is able to recognize a human activity (e.g., walking, cooking etc.). The accuracy of the classification algorithm depends heavily on the extracted features, while the feature extraction process is time consuming.

Deep Learning (DL) [10] can provide a solution to this problem. DL is a branch of Machine Learning (ML), in particular, Artificial Neural Networks (ANNs), and has the ability to automatically extract features. What is more, previous implementations of DL approaches to computer tasks such as computer vision [11], speech recognition [12] and natural language processing [13], have outperformed past techniques based on HCFs. As a result, state-of-the-art HAR methods (e.g., [14][15]) are based on DL. In this paper, we adopt a DL approach, to process and analyze multimodal sensor data produced by mobile motion sensors. In our proposal, we apply late sensor fusion (2D convolution), to recognize the patterns of human activities. More specifically, this is the first 2D convolution on multimodal raw motion sensor data (to the best of our knowledge). The contributions and innovations of our proposal can be summarized in the following:

- Proves that late sensor fusion is more effective in Deep Convolutional Neural Networks.

- Applies a 2D convolution to vertically stacked motion sensor data.

- Utilizes global average pooling over feature maps in the classification layer, instead of traditional fully connected layers.

- Outperforms other state-of-the-art deep learning techniques in HAR.

The rest of the paper is organized as follows: In Section II, an overview of the state-of-the-art in deep learning approaches to HAR is presented. In the next Section, the proposed architecture of the deep convolutional neural network is explained. Section IV describes the experimental set up, while Section V presents the results of the proposed method using two public datasets. Finally, Section VI concludes the paper and proposes future work directions.

## II. State-of-the-art

After the raw data collection, a typical human activity recognition system deploys tools and techniques for preprocessing, segmentation, feature extraction and classification [16]. However, DL approaches to HAR have revealed that the step of feature extraction is included in the DL algorithm [17]. One of the greatest advantages of Deep Neural Networks (DNNs), is their ability to extract their own features, which manage to express complex relations/patterns between the data [18]. Thus, DL approaches to HAR are considered to be the state-of-the-art, and are categorized as follows: a) *Autoencoders*, b) *Convolutional Neural Networks*, c) *Deep learning on spectrograms* and d) *Convolutional Recurrent Neural Networks*. In the following paragraphs, a summary of existing DL work towards HAR, in all four categories will be presented.

### A. Autoencoders

Autoencoders are a specific field of ANNs, based on unsupervised learning (i.e., machine learning based on unlabeled data), where an ANN consisting of one hidden layer, tries to produce as output the input values. Many stacked Autoencoders form a DNN. In [19] an Autoencoder technique was used to extract key features for HAR. In particular, the authors used Restricted Bozltman Machines (RBMs) [20], which are a particular form of log-linear Markov Random Fields (i.e., a non-directed probabilistic graphical model) and has been applied successfully for dimensionality reduction in computer vision [21]. Another autoencoder approach was proposed in [22], where C. Vollmer et al use a Sparse Autoencoder [23] for extracting features. Sparse Autoencoders have been successfully applied to computer vision problems, such as medical image analysis [24][25]. However, Autoencoders are fully connected DNN models, and as a result, do not manage to capture the local dependencies of the time-series sensor data [26][27].

### B. Convolutional Neural Networks

Yann LeCun et al. introduced LeNet, a Convolutional Neural Network (CNN) in [28]. Based on the mathematical operation of convolution (i.e., the combination of two functions to form a third one), LeNet managed to outperform the other classification algorithms in recognizing hand-written digits. However, ConvNets (Convolutional Networks) drew public attention almost 15 years later, where the deep ConvNet of A. Krizhevsky et al. [11], called AlexNet, surpassed the performance of the runner-up algorithm by almost 5%, using the ImageNet dataset [29]. Since then, DL has become the state-of-the-art method in various computer tasks (e.g., natural language processing).

The first CCN approach to HAR was introduced in [30]. The authors used as input a 1D array representation of the motion signals, unlike image analysis that uses a 2D array of pixels. In this way, the signals are stacked into channels (channel-based stacking). For example, a tri-axial accelerometer sensor produces 3 channels (X, Y, Z axis), similarly to colored images (Red, Blue, Green channels). As a result, the convolution operation is applied to each signal individually. After two successive *Convolutional* and *Pooling* (*ConvPool*) operations ([31]), they concatenated the channels into a 1D array and applied a Multilayer Perceptron (MLP), otherwise a Dense layer, to do the classification.

Similarly, in [27] they propose the same architecture, but they use a more shallow CNN (only one *ConvPool* operation), having as input 3D acceleration time series. The results they acquired showed that CNN approach outperforms the Autoencoders. In addition to this, Ronao et al. [14][32][33] and J. B. Yang et al. [34] propose a deeper CNN (3 *ConvPool* layers) approach to HAR, relying on a dataset with tri-axial gyroscope and accelerometer data. Moreover, introducing additional information as input, by adding the Fourier transformation, acquired, almost, a 1% higher accuracy [14]. It should be noted that all the aforementioned CNN approaches fuse the motion data in the first hidden layer and use as final hidden layer an MLP (i.e., similarly to Autoencoders some local time-dependent patterns are not discovered). Thus, we will refer to them for the rest of the paper using the abbreviation CNN-EF (Early Fusion).

### C. Deep learning on spectrogram

An interesting approach which is applied in audio signals, is that of converting the input sensor signal to spectrogram, during feature extraction step, providing a representation of the signal as a function of frequency and time. Afterwards, the spectrogram image feeds a DNN, similarly to the image analysis process.

Alsheikh et al. [35] adopt a hybrid approach of Deep Learning and Hidden Markov Models (DL-HMM) for sequential activity recognition. Specifically, the tri-axial accelerometer signal is translated into spectrogram and afterwards a RBM is applied. Furthermore, a non-mandatory HMM step, which has as input the emission probabilities out of the DNN, is used for modeling temporal patterns in activities. A more sophisticated technique is adopted in [36], where the CNN has as input an *activity image*. According to the authors, the raw tri-axial accelerometer and gyroscope signals are stacked row-by-row into a signal image, based on an algorithm. Subsequently a 2D Discrete Fourier Transform (DFT) is applied to the *signal image* and its magnitude is represents the *activity image*. This way, signal sequences are adjacent to other sequences, enabling the DNN to extract hidden correlations between neighboring signals. However, it should be noted that, conversion of time-series data into the frequency domain is not as effective in HAR as in audio classification, and time statistical features have proven to be more essential [9].

### D. Convolutional Recurrent Neural Networks

Recurrent Neural Networks (RNNs) [37] are a family of neural networks for processing a sequence of values. As a result, RNNs are applied broadly to time-series data. Moreover, because in case of sequential data, a value $xi$ depends on a set of previous $n$ values $\{x_{i-1}, x_{i-2}, \ldots, x_{i-n}\}$ and on a set of next n values $\{x_{i+1}, x_{i+2}, \ldots, x_{i+n}\}$, a mechanism named LSTM (Long Short-Term Memory) [38] is applied to enhance the memory of the network.

A hybrid approach called Convolutional LSTM is presented in [15]. This network consists of four consecutive 1D convolution operations, which have as input vertically stacked motion signals, where the output of the last feeds a LSTM layer. Afterwards, a final LSTM layer is used to predict the class of the activity performed. As a result, the signals are not fused during the convolutions, but they are convolved using the same filter. This RNN approach was applied on two HAR datasets and managed to model temporal dependencies more effectively than a conventional ConvNet.

Furthermore, the same authors, using the same network architecture, studied the possibilities of transfer learning (e.g., transferring trained filters) in CNNs layer by layer for activity recognition based on wearable sensors [39]. The experimental results show that the performance of the model for the same application domain was not affected by transferring features of the first layer, while there was an improvement in training time (~17% reduction). However, the accuracy of the algorithm was significantly reduced after applying transfer learning between different applications domains, between sensor locations and between sensor modalities.

Following the heuristic that the motion signals are correlated with each other, which is denoted in [36], in this paper we propose a different representation of time-series sensor data (vertical stacking) that applies a late sensor fusion and allows a 2D convolution operation over them.

## III. PERCEPTIONNET

In this paper, we introduce the concept of applying a late 2D convolution on HAR data in an attempt to avoid overfitting and discover more general activity patterns, emanating from the cross-correlation between high-level features of the motion signals. We developed a deep CNN model, named PerceptionNet, having as input vertically stacked motion signals in order to exploit the semantics and the grid-like topology of the input data, in contrast with conventional ANNs. The intuition for applying late sensor fusion, the components/layers of PerceptionNet, and the selected optimizer are described below.

### A. Convolutional Layer

The convolution operation manages to obtain a less noisy estimate of a sensor's measurements, by averaging them. Because of the fact that some measurements should contribute more in the average, the sensors measurements are convolved with a weighting function $w$ [31]. Consequently, in our case the input, which is the preprocessed and segmented motion signal, is combined with the filters (weights), which are trained in order to discover the most suitable patterns (e.g., peaks in the signal). Moreover, each filter is replicated across the entire signal. As a result, the replicated units share the same weight vector and bias and form a *feature map* (or *activation map*), which is the product of several convolutions in parallel of the signal and the *filters*.

In other words, all the signal values in a convolutional layer respond to the same feature within a specific receptive field [40]. This iteration over all the units allows for motion features to be detected regardless of their position in the sensor signal (translation invariance property). In particular, the $i$th product element of a discrete 1D convolution between input array $x$ and a 1D filter $w$ equals:

$$c_i^{l,q} = b^{l,q} + \sum_{d=1}^{D} w_d^{l,q} x_{i+d-1}^{l-1,q} \tag{1}$$

where $l$ is the layer index, $q$ is the activation map index, $D$ is the total width of the filter $w$, and $b$ is the bias term. However, in case there are more than one channels, (i.e., the sensor signals are stacked by the channel axis), the $i$th product elements ($c_i$) of the sensor signals are added, producing a new element ($c_{i,j}$):

$$c_{i,j}^{l,q} = b^{l,q} + \sum_{h=1}^{H} \sum_{d=1}^{D} w_{d,h}^{l,q} x_{i+d-1}^{l-1,q} \tag{2}$$

where $h$ is the channel index. This way, the translation invariance property is lost, since the specific receptive fields of the signals may not be correlated. In addition to this, motion signals produced by low-cost, not well-calibrated sensors, suffer from sampling rate instability (regularity of the timespan between successive measurements) [41].

In order to understand this issue better, imagine a picture showing a cat, where the R, G, B channels are not stacked correctly (e.g., the nose in the red channel matches the eye in the blue channel, and the mouth in the green channel). If a 2D convolution was applied to this picture, it would detect in each channel different edges, and, as result, by adding them it would extract features only with respect to this particular R, G, B topology, resulting to overfitting. However, if the CNN model identified the low-level features (e.g., edges), and the mid-level features (e.g., nose) for each channel separately, it would have the capability to "see the whole picture" (i.e., perceive the high-level features) and generalize what it learned.

### B. Architecture

The architecture of PerceptionNet, illustrated in Fig. 1, consists of the following layers:

Layer 1: 48 1D convolutional filters with a size of (1,15), i.e., $W^1$ has the shape (1, 15, 1, 48). This is followed by a ReLU [11] activation function, a (1,2) strided 1D max-pooling operation and a dropout [42] probability equal to 0.4.

Layer 2: 96 1D convolutional filters with a size of (1,15), i.e., $W^2$ has the shape (1, 15, 48, 96). This is followed by a ReLU activation function, a (1,2) strided 1D max-pooling operation and a dropout probability equal to 0.4.

Layer 3: 96 2D convolutional filters with a size of (3,15) and a stride of (3,1), i.e., $W^3$ has the shape (3, 15, 96, 96). This is followed by a ReLU activation function, a global average-pooling operation [43] and a dropout probability equal to 0.4.

Layer 4: 10 output units, i.e., $W^4$ has the shape (96, 6), followed by a softmax activation function.

$$g_t = (1 - \rho)L'(\theta_t)^2 + \rho g_{t-1} \tag{3}$$

where $g_0 = 0$ and $s_0 = 0$. The term $s_t$ denotes the 2nd moment of $\Delta \theta_t^2$ for updating the parameters:

$$\Delta \theta_t = -\frac{\sqrt{s_{t-1} + \varepsilon}}{\sqrt{g_t + \varepsilon}} L'(\theta_t) \tag{4}$$

$$s_t = (1 - \rho)\Delta \theta_t^2 + \rho s_{t-1} \tag{5}$$

$$\theta_{t+1} = \theta_t + \Delta \theta_t \tag{6}$$

## IV. EXPERIMENTAL SET UP

The experiments were executed on a computer workstation equipped with an NVIDIA GTX Titan X GPU, featuring 12 gigabytes RAM, 3072 CUDA cores, and a bandwidth of 336.5 GB/s. We used Python as programming language, and specifically the Numpy library for matrix multiplications, data preprocessing and segmentation, the scikit-learn library for implementing the t-SNE algorithm, and the Keras high-level neural networks library using as backend the Theano library. In order to accelerate the tensor multiplications, we used CUDA Toolkit in support with the cuDNN, which is the NVIDIA GPU-accelerated library for deep neural networks. The software is installed on a 16.04 Ubuntu Linux operating system.

### A. Datasets

We evaluated PerceptionNet on two public available HAR datasets, UCL [45] and PAMAP2 [46]. The first one, was used to tune the hyper-parameters and we compared, afterwards, our CNN's performance against the state-of-the-art approaches: a) CNN-EF [14], b) Convolutional LSTM [15], c) CNN on spectrograms [35], and d) SVM method based on HCF [45]. Finally, in order to test the general applicability of our approach we used the same CNN architecture on the PAMAP2 dataset and compared it with the conventional CNN-EF [14] and the Convolutional LSTM [15] methods.

#### 1) UCL

The UCL HAR dataset consists of tri-axial accelerometer and of tri-axial gyroscope sensor data, collected by a waist-mounted smartphone (Samsung Galaxy S II smartphone). A group of 30 volunteers, with ages ranging from 19 to 48 years, executed six daily activities (standing, sitting, laying down, walking, walking downstairs and upstairs). The mobile sensors produced 3-axial linear acceleration and 3-axial angular velocity data with a sampling rate of 50 Hz and were segmented into time windows of 128 values (2.56 sec), having a 50% overlap. Furthermore, the dataset is separated into train data.

The obtained dataset contains 10,299 samples, which are partitioned into two sets, where 70% of the volunteers (21

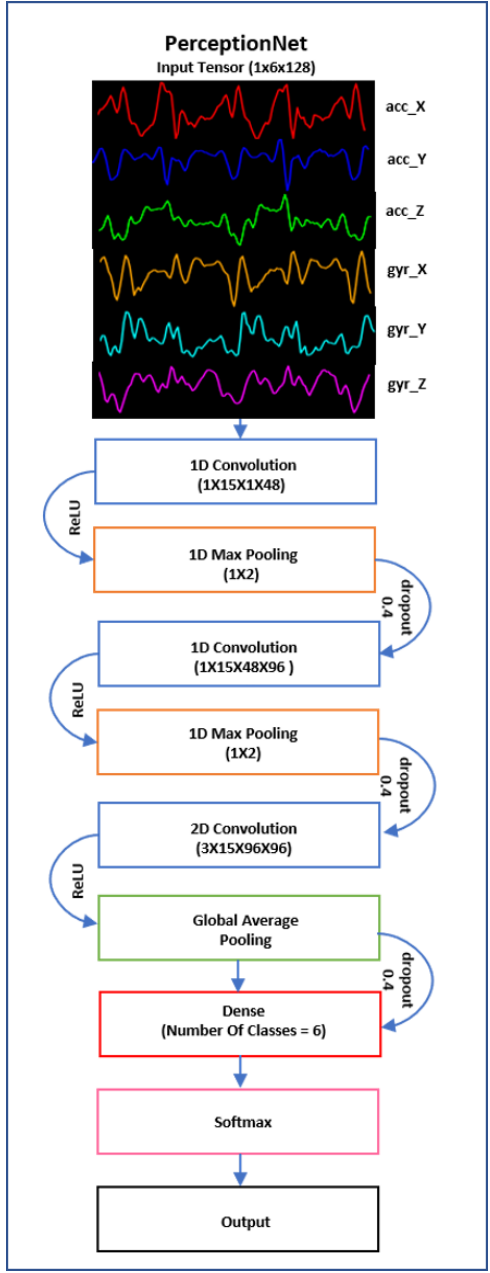

Fig. 1: PerceptionNet architecture

### C. Adadelta optimizer

We selected the *adadelta* [44] as optimizer in our network. The reason for this was because *adadelta* adapts dynamically over iterations, does not need manual tuning of the learning rate and appears robustness to noisy gradient information, different model architecture choices, various data modalities and selection of hyper-parameters. According to [44], let $L'(\theta_t)$ be the first derivative of the loss function $f$ with respect to the parameters $\theta$ at time step $t$. Here $g_t$ is called the second order moment of $L(\theta_t)^2$. Given a decay term $\rho$ and an offset $\varepsilon$ we perform the following updates:

volunteers) was selected for generating the training data (7,352 samples) and 30% (9 volunteers) the test data (2,947 samples). Moreover, following the results in [47] where it was shown that subject independent validation techniques should be applied for the evaluation of activity monitoring systems to tune the hyper-parameters, during the validation phase we followed a *Leave-3-Subject-Out* approach. Concretely, the samples of 3 volunteers (27, 29 and 30) were used as validation set, which are equal to 15% of the training set.

Finally, we normalized each sensor's values ($x_i$) by subtracting the mean and dividing by the standard deviation:

$$z_i = \frac{x_i - \mu_i}{\sigma_i} \tag{7}$$

*2) PAMAP2*

The PAMAP2 HAR dataset contains 12 lifestyle activities (such as walking, cycling, ironing, etc.) from 9 participants wearing 3 Colibri wireless inertial measurement units (IMU) and a heart rate monitor. The 3 IMUs had a sampling frequency of 100Hz, were placed on the dominant arm, on the chest and on the dominant side's ankle, and produced tri-axial accelerometer, gyroscope and magnetometer data. In order to obtain the same sampling rate with the UCL dataset and the same sensor signals, we downsampled the PAMAP2 dataset to 50Hz and selected only the accelerometer and gyroscope data. The resulting dataset had 18 dimensions, with the same time window (2.56 sec) and overlap (50%) as the UCL dataset.

Since the data were collected by only 9 participants and for the reason that only 4 subjects (1, 2, 5, and 8) have enough samples of all the activities, we selected a *Leave-1-Subject-Out* approach, for the test and the validation set. More specific, the samples of subject 1 were used for the test set and the samples of subject 5 for the validation set. The training set contained 13,980, the test set 2,453 and the validation set 2,688 samples. The PAMAP2 samples were, also, normalized using (7).

*B. Performance metrics*

We used precision, recall, weighted (w) F1-score (otherwise F-measure), and accuracy as performance measures. It should be noted that accuracy, in contrast with the other 3 metrics, takes only into account the total number of samples and not class imbalance. On the other hand, precision, recall, and wF1-score consider total number of samples for each class separately. The above metrics are described as:

$$accuracy = \frac{1}{N} \sum_{i=1}^{N} (TP_i + TN_i) \tag{8}$$

$$precision = \frac{1}{N} \sum_{i=1}^{N} \left( \frac{TP_i}{TP_i + FP_i} \right) \tag{9}$$

$$recall = \frac{1}{N} \sum_{i=1}^{N} \left( \frac{TP_i}{TP_i + FN_i} \right) \tag{10}$$

$$F_1 = \sum_{i} 2 * w_i \frac{precision_i * recall_i}{precision_i + recall_i} \tag{11}$$

where TP, TN, FP, FN represent the true positive, true negative, false positive and false negative predictions respectively. It should be noted that in a multiclassification problem, the recall, precision and wF1-score metrics iterate over all the classes by selecting the samples belonging to one of them as positive (class of interest), while they consider the rest samples as negative (rest of the classes).

Moreover, we selected the confusion matrix as a visualization of the classification performance of PerceptionNet. The confusion matrix is easy to be interpreted; it shows where the classification algorithm "confused" a class with another one (i.e., it predicted lying activity instead of standing activity). In mathematical terms, the confusion matrix can be described by $M_{ij}$, with $i$ denoting the "actual" classes and $j$ the "predicted" classes. By summing all entries in the row $i$ of the matrix it shows the total number of the samples annotated as activity $i$, while by summing all entries in the column $j$ of the matrix it shows the total number of the samples predicted as activity $j$.

V. RESULTS

*A. Validation phase*

Before, testing our approach, we used the validation set of UCL to tune its hyper-parameters. Table I contains all the hyper-parameters and their possible values. Since it is time consuming to train a deep CNN model, and because the UCL dataset was thoroughly examined in [14], we did not evaluate exhaustively all potential combinations of the hyper-parameters. Thus, we based the selection of the most promising model on the most significant factors (filter shape, dropout probability, layer of 2D convolution, and use of dense layer or global average pooling) that differentiate our method from that described in [14].

Since the convergence and the prediction accuracy of a Deep Neural Network depend a lot on the weight initialization [48], in order to obtain more representative results for each hyper-parameter, we ran the experiments 10 times. The mean value of each different hyper-parameters model was used as selection criterion. It should be noted that we selected a variation (i.e., the random numbers were sampled from a uniform instead of a normal distribution) of the weight initialization introduced by K. He et al. [49], whose upper and lower thresholds are given by:

$$w_i = \pm \sqrt{\frac{2}{N_{in}}} \tag{12}$$

TABLE I. EXPERIMENTAL SET-UP FOR TUNING HYPER-PARAMETERS

| Symbol | Parameter | Values |
|---|---|---|
| - | batch size | 64 |
| $\alpha$ | learning rate | 1.0 |
| $\rho$ | rho | 0.95 |
| $\varepsilon$ | epsilon | 1e-08 |
| - | number of channels | 1 |
| - | input height | 6 |
| - | input width | 128 |
| - | number of convolutional layers | 3 |
| - | 1D convolution size | 1X5-1X17 |
| - | 2D convolution size | 3X15 |
| - | 2D strides size | 3X1 |
| - | 1D max pooling size | 1X2 |
| - | 1D max pooling stride | 1X2 |
| - | dropout | 0-0.7 |
| - | activation map channels | 32-192 |
| - | dense layer size | 0-1500 |
| - | 2D convolutional layer | 1-3 |
| - | maximum epochs | 2000 |
| - | early stopping criterion epochs | 100 |

where $N_{in}$ represents the total numbers of neurons that outputs the *(i − 1)-th* layer, which are the input regarding the next layer. The adadelta optimizer had the following hyper-parameters: learning rate equal to 1, $\rho$ equal to 0.95, and $\varepsilon$ equal to 1e-08. Moreover, we set the batch size equal to 64 and the minimum number of epochs to 2,000, but the training procedure was automatically terminated if the best training accuracy had not improved after 100 epochs. The model that achieved the lowest error rate on the validation set was saved, and its filters were used to obtain the accuracy of the model on the test set.

Before, testing our approach, we used the validation set of UCL not only to tune its hyper-parameters (Table I), but to show also that fusing sensor signal in the latest convolutional layer is more effective. Thus, we applied the 2D convolutions on three different convolutional layers, each of them having filter size 3x15 and stride equal to (3,1), for the vertical and the horizontal axis respectively. Fig. 2 shows that applying the 2D convolution on the last convolutional layer increases the accuracy of the model. The increased performance of applying a late 2D convolution is, also, illustrated in Fig. 3, which shows that by applying the t-SNE algorithm [50] after the last convolutional operation most of the instances of the six activity classes are easily categorized.

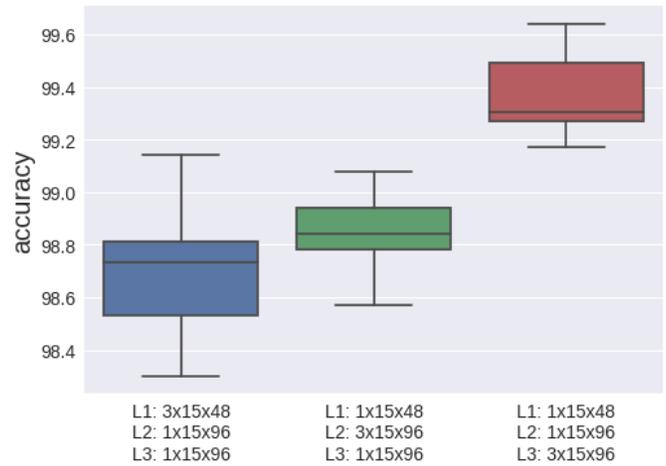

Fig. 2. Accuracy results of 2D convolutions on the 1st, 2nd and 3rd convolutional layer of the UCL validation set.

Moreover, we examined the concept of having a dense layer, as last hidden layer and that of using a global average pooling or a max average pooling operation instead of max pooling operation. As it is shown in Fig. 4, we obtained the highest mean accuracy with a global average pooling operation (99.38%).

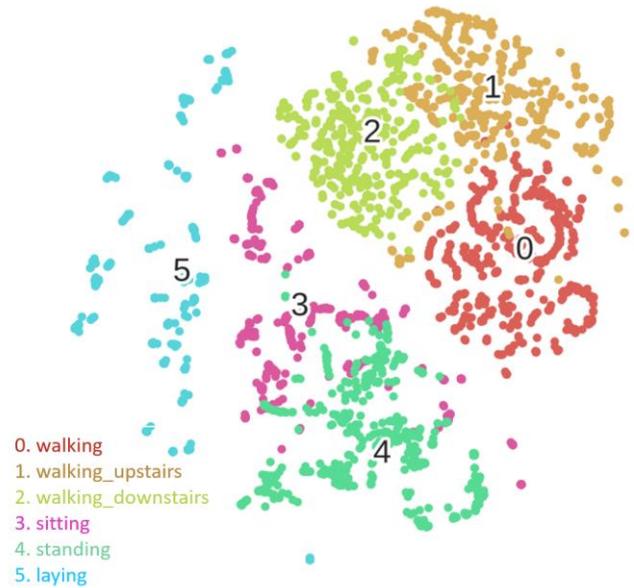

0. walking
1. walking_upstairs
2. walking_downstairs
3. sitting
4. standing
5. laying

Fig. 3. t-SNE visualization of the test set's last hidden layer representations in PerceptionNet for six activity classes.

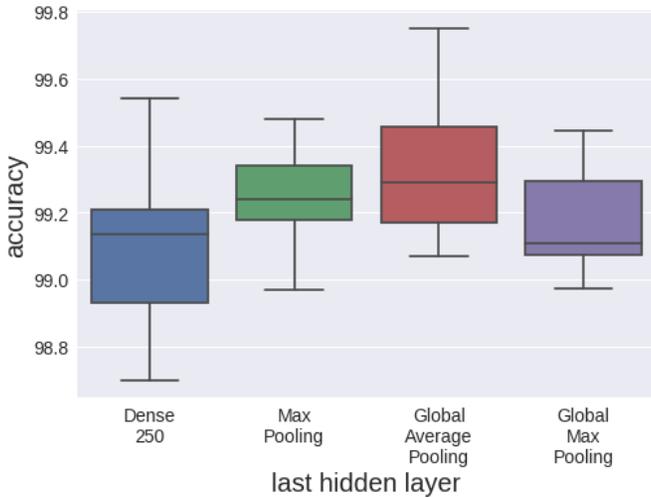

Fig. 4. Accuracy results based on different last hidden layers of the UCL validation set.

## B. UCL Test phase

The results we obtained using PerceptionNet model on the test set are presented in Fig. 5, and have a range from 0.9620 to 0.9752. After obtaining the results, we developed an CNN ensemble, based on probability voting (e.g., the probability of the 10 runs of our model were added and divided afterwards by 10). Fig. 6 presents the CNN ensemble. The best epoch on the validation data achieved an 0.9725 accuracy on the test data, which is about 2.5% higher than the ConvNet described in [14]. Table II compares the accuracy obtained from the PerceptionNet against those obtained by state-of-the-art models, whose results are reported in the literature, and our implementation of the Convolutional LSTM [15].

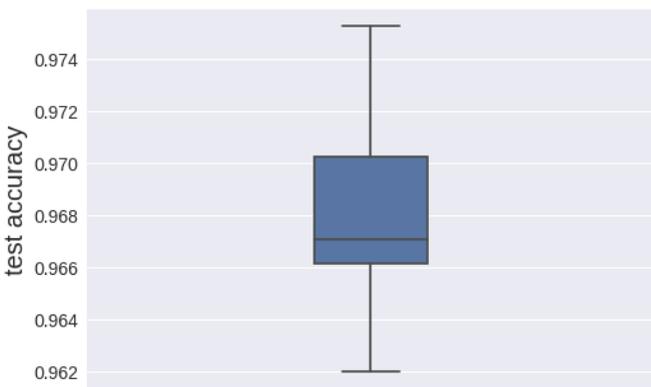

Fig. 5. Test accuracies of PerceptionNet on the UCL set.

```
function probability_ensemble(trained_cnn_models, num_classes, test_data):
    foreach i belonging in test_data
        foreach j belonging in num_classes
            foreach k belonging in trained_cnn_models
                total_preds(i, j, k) ← predict(i, j, k)
        mean_total_preds ← mean_{trained_cnn_models}(total_preds)
        ensemble_preds ← argmax_{trained_cnn_models}(mean_total_preds)
    return ensemble_preds
```

Fig. 6. Pseudocode for the average probability ensemble.

TABLE II
COMPARISON OF PERCEPTIONNET TO OTHER STATE-OF-THE-ART METHODS.

| Method | Accuracy on test data |
|---|---|
| CNN-EF [14] | 94.79% |
| CNN-EF + FFT features [14] | 95.75% |
| SVM on HCF [45] | 96.00% |
| CNN on spectrogram [35] | 95.18% |
| Convolutional LSTM [15] | 92.59% |
| PerceptionNet | **97.25%** |

The accuracy per subject of our 2D ConvNet is presented in Table III, while Fig. 7 presents the confusion matrix (precision: 0.9731, recall: 0.9725, and wF1-score: 0.9724), which reveals the difficulty of distinguishing standing from sitting and the reverse.

TABLE III
ACCURACY OF PERCEPTIONNET ON EACH SUBJECT OF THE TEST SET

| Subject | Accuracy on test data |
|---|---|
| 2 | 98.01% |
| 4 | 98.10% |
| 9 | 90.95% |
| 10 | 92.49% |
| 12 | 96.24% |
| 13 | 100.0% |
| 18 | 99.45% |
| 20 | 98.01% |
| 24 | 100.0% |

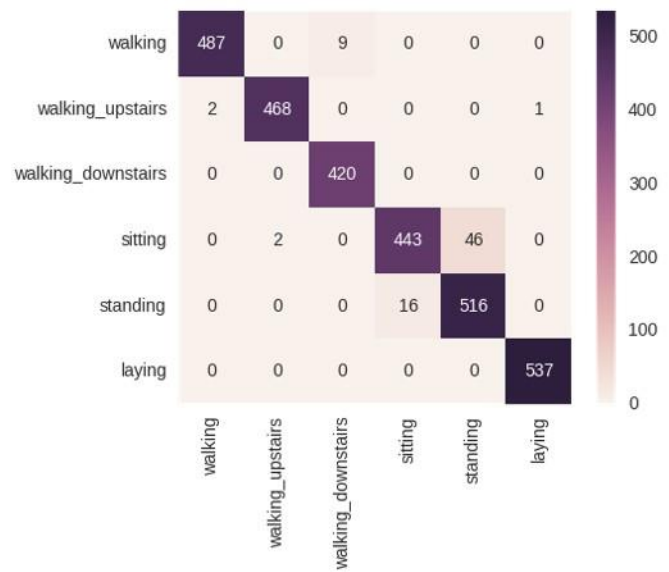

Fig. 7. Confusion matrix of PerceptionNet on the UCL test data.

*C. PAMAP2 Test phase*

PAMAP2 dataset was selected as a second dataset, in order to examine the general applicability of our approach. Table IV presents the precision, the recall, the wF1-score and the accuracy of PerceptionNet ensemble on the test data, compared with the ensembles of the conventional CNN (channel-based stacking) and the Convolutional LSTM methods. PerceptionNet achieved almost 2% higher accuracy form the Convolutional LSTM and 4% higher than the CNN-EF approach.

Fig. 8 shows the confusion matrix of our model on the PAMAP2 test data. Not surprisingly, the model again struggles to distinguish the sitting activity from the standing activity. The authors of [51][52] argue that this misclassification is a common problem, and an extra IMU on the thigh would be a solution. Finally, it should be noted that the ironing class had very high recall (0.9913), but very low precision (0.6930) indicating a large number of False Positives.

Future steps towards improving the model's performance include the use of Capsule Networks and the Dynamic Routing [53] mechanism to achieve a more efficient sensor fusion. Moreover, in order to reduce the training time without affecting negatively the PerceptionNet's performance, transfer learning techniques should be studied. Finally, the features that are extracted after the global average pooling layer can be used as embedding and one-shot learning techniques should be investigated in the future.

ACKNOWLEDGMENT

This work is funded by the European Commission under project TRILLION, grant number H2020-FCT-2014, REA grant agreement n° [653256]. Moreover, we gratefully acknowledge the support of NVIDIA Corporation with the donation of the Titan-X GPU used for this research.

TABLE IV
COMPARISON OF PERCEPTIONNET TO OTHER STATE-OF-THE-ART METHODS ON THE PAMAP2 TEST DATA.

| Method | Precision | Recall | wF1-score | Accuracy |
|---|---|---|---|---|
| CNN-EF [14] | 85.51% | 84.53% | 84.57% | 84.53% |
| Convolutional LSTM [15] | 87.75% | 86.78% | 86.83% | 86.78% |
| PerceptionNet | **89.76%** | **88.57%** | **88.74%** | **88.56%** |

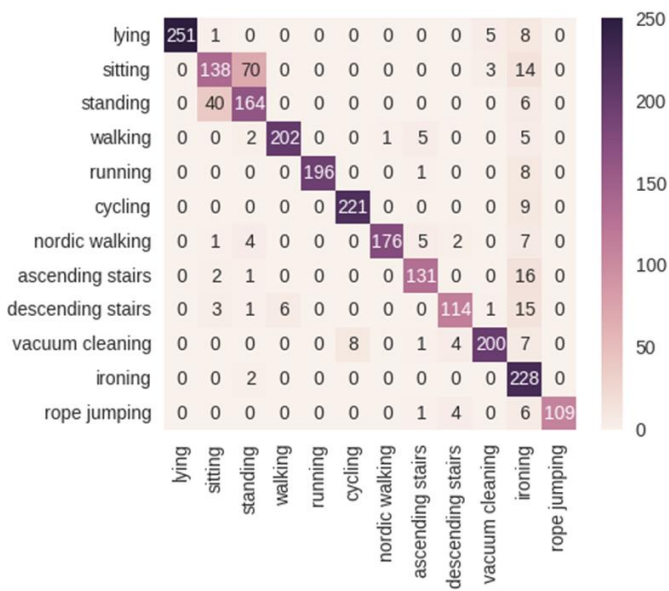

Fig. 8. Confusion matrix of PerceptionNet on the PAMAP2 test data.

VI. CONCLUSION

In this paper, we propose a deep Convolutional Neural Network (CNN) for human activity recognition (HAR) that performs a 2D convolution on the last convolutional layer.